\documentclass[journal]{IEEEtran}
\usepackage{cite}
\usepackage[utf8]{inputenc}
\usepackage{fancyhdr} 
\usepackage{tikz}
\usetikzlibrary{shapes,backgrounds}
\usepackage{verbatim}
\usepackage{url}
\usepackage{enumitem}
\usepackage{dirtytalk}
\usepackage{amsmath,amssymb,amsfonts}
\usepackage{xfrac}
\usepackage{algorithmic}
\usepackage{graphicx}
\usepackage{textcomp}
\usepackage{xcolor}
\usepackage{color,soul}
\usepackage{scalerel}
\usepackage{csquotes}
\usepackage{subcaption}
\usepackage{multirow}
\usepackage{tikz}
\usetikzlibrary{svg.path}
\usepackage[ruled,vlined]{algorithm2e}
\usepackage{nomencl}
\makenomenclature

\usetikzlibrary{svg.path}

\definecolor{googleblue}{rgb}{0.259,0.522,0.957}
\definecolor{googlegreen}{rgb}{0.060,0.620,0.350}
\definecolor{googlered}{rgb}{0.859,0.267,0.220}
\definecolor{bleudefrance}{rgb}{0.19,0.55,0.91}

\definecolor{orcidlogocol}{HTML}{A6CE39}
\tikzset{
  orcidlogo/.pic={
    \fill[orcidlogocol] svg{M256,128c0,70.7-57.3,128-128,128C57.3,256,0,198.7,0,128C0,57.3,57.3,0,128,0C198.7,0,256,57.3,256,128z};
    \fill[white] svg{M86.3,186.2H70.9V79.1h15.4v48.4V186.2z}
                 svg{M108.9,79.1h41.6c39.6,0,57,28.3,57,53.6c0,27.5-21.5,53.6-56.8,53.6h-41.8V79.1z M124.3,172.4h24.5c34.9,0,42.9-26.5,42.9-39.7c0-21.5-13.7-39.7-43.7-39.7h-23.7V172.4z}
                 svg{M88.7,56.8c0,5.5-4.5,10.1-10.1,10.1c-5.6,0-10.1-4.6-10.1-10.1c0-5.6,4.5-10.1,10.1-10.1C84.2,46.7,88.7,51.3,88.7,56.8z};
  }
}

\newcommand\orcidicon[1]{\href{https://orcid.org/#1}{\mbox{\scalerel*{
\begin{tikzpicture}[yscale=-1,transform shape]
\pic{orcidlogo};
\end{tikzpicture}
}{|}}}}

\usepackage[bookmarks=false]{hyperref} 
\hypersetup{
    colorlinks=true,
}

\begin{document}
\raggedbottom

\twocolumn
\setcounter{page}{1}

\title{SymboSLAM: Semantic Map Generation in a Multi-Agent System}

\author{
        \IEEEauthorblockN{Brandon C. Colelough,~$^{\orcidicon{0000-0001-8389-3403}}$}
        
        \IEEEauthorblockA{School of Engineering and Information Technology, University of New South Wales, Australia}
}

\maketitle

\begin{abstract}

Sub-symbolic artificial intelligence methods dominate the fields of environment-type classification and Simultaneous Localisation and Mapping. However, a significant area overlooked within these fields is solution transparency for the human-machine interaction space, as the sub-symbolic methods employed for map generation do not account for the explainability of the solutions generated. This paper proposes a novel approach to environment-type classification through Symbolic Simultaneous Localisation and Mapping, \emph{SymboSLAM}, to bridge the explainability gap. Our method for environment-type classification observes ontological reasoning used to synthesise the context of an environment through the features found within. We achieve explainability within the model by presenting operators with environment-type classifications overlayed by a semantically labelled occupancy map of landmarks and features. We evaluate SymboSLAM with ground-truth maps of the Canberra region, demonstrating method effectiveness. We assessed the system through both simulations and real-world trials.

\end{abstract}


\section{Introduction} \label{sec:intro}

An environment is a \enquote{broadly defined term} that refers to the \enquote{natural or anthropogenic systems which can surround a living or non-living entity}~\cite{ENV_ONTO}. The symbolic constructs attributed to environments are \enquote{created by human acts of conferring meaning to nature and the environment}~\cite{greider1994landscapes}. Thus, environments may be represented by symbolic concepts and described through their physical attributes~\cite{greider1994landscapes, ENV_ONTO}. This study focuses on assigning symbolic concepts to areas of an environment by examining the physical objects found within. Making this link between an object and an environment type is a problem that requires some level of human understanding present within the methodologies for classification, as the environment labels are complex societal constructs. Encoding this knowledge within an architecture requires a system that can understand the nuances and intricacies of the human world and the relationships people draw between environmental concepts and the objects associated with them. These manifest as challenges in modelling an environment for the purpose of classification, raising questions such as:

\begin{enumerate}
    \item How can a symbolic representation of an environment be used to classify the environment type?
    \item How can an environment be transformed into a symbolic representation?
\end{enumerate}

Symbolic Artificial Intelligence (AI) is a field of study that offers a solution to question one. It incorporates domain-level expertise within an architecture to enable higher-order reasoning about concepts. An ontology is an instrument within the field of symbolic reasoning which includes the properties and relations of objects and ideas within a specific subject area~\cite{jean-baptiste_2021}. It posits a viable solution to link the concepts of objects and environment types. It is evident from the literature presently available within symbolic reasoning and remote sensing fields that the information required for enabling environment-type classification is not readily available or easily obtainable. As such, a method to extract information is necessary to allow higher-order classification of the environment. Simultaneous Localisation and Mapping (SLAM) offers a solution having excellent applicability over a range of domains for which many systems have already demonstrated promising results~\cite{Kimera_V2, magic, Hierarchical_prob_mapping}. Joo et al. demonstrated that it is possible to generate semantic understanding from defined spatial relations~\cite{TOSM}. The incorporation of semantic knowledge to features in the environment offered a new method to describe SLAM situations transparently. 
The overarching motivation to research environment-type classification through symbolic reasoning was to transparently describe complex environments with a simple graphical representation. Transparency within a system can be achieved through the tenets of \enquote{interpretability, explainability and predictability} and is the \enquote{overarching concept} required for a human-machined teaming environment trust architecture~\cite{swarm_ontology}. Using an ontology for symbolic reasoning offers the opportunity to increase the explainability and interpretability of the classification result, enabling transparency to be maintained within the human-machine teamed environment. Thus, transparency within a system may be achieved by conducting high-level environment classification through the contextualisation of objects found within and displaying these classifications as a 2D map. Furthermore, an architecture that utilises an ontology as the symbolic reasoning component will be capable of encoding human constructs into the architecture’s knowledge base to ensure the architecture converges upon a solution. The identified gap in the literature leads our research to answer the following overarching question; \emph{can a symbolic reasoning approach for multi-agent SLAM increase the meaning for 2D maps?} The objectives of this research are:

\begin{enumerate}[label=RO\arabic*]
    \item Determine whether intelligent edge agents on a Multi-Agent System (MAS) can extract environmental features and share information with a centralised control agent.
    \item Determine if a centralised control structure with intelligent MAS agents can conduct SLAM.  
    \item Determine if a symbolic representation of the environment is suitable to build classified environment-type 2D maps.
\end{enumerate}

The main contribution of this study is a transparent method for environment-type classification, achieved through explainability within the map generation process. This study presents system explainability for environment-type reasoning through semantically labelled landmarks as an occupancy map. Furthermore, this study provides ontological reasoning of the semantically labelled landmarks as a method to conduct environment-type classification.  


The remainder of this paper is organised as follows. First, in Section~\ref{sec:bg}, we review methods for feature extraction and spatial recognition on a MAS and symbolic reasoning through ontological design. The problem space is then constructed to describe both the project sub-tasks and how they relate to the overall aim of the proposed architecture. Next, Section~\ref{sec:methodology} introduces the proposed solution to the main research question and research sub-objectives, with development steps in the appendix. Following this within Section~\ref{sec:Eval}, the metrics for success across all project contributions are introduced, and the method to evaluate the proposed SymboSLAM architecture is discussed. Finally, in Section~\ref{sec:Results} the simulated and measured experiments conducted with the SymboSLAM architecture are presented alongside the results gathered with supplementary information provided in the appendix and externally. We then conclude this paper with a discussion on the success of the proposed architecture in Section~\ref{sec:Discussion}, followed by the proposal of further research questions for further investigation in Section~\ref{sec:fut_work}.

\section{Background}\label{sec:bg}
\subsection{Environment Type Classification}
Environment type classification is a field of study presently dominated by the implementation of remote sensing applications utilised as the primary source of information. These remote sensing techniques deal with \enquote{high-resolution geospatial data} to \enquote{find the land cover classes} and are broadly sorted into supervised, unsupervised and object-orientated classification techniques~\cite{8474881}. Semi-supervised classification techniques of hyperspectral images are used by Wang et al.~\cite{8071701} to classify non-urban regions of China, with a reported accuracy above 90\%. Zhang et al.~\cite{6927922} compare land cover classification methods in an arid/semi-arid environment that demonstrates the ability of object-orientated classification techniques to differentiate between populated and non-populated areas. However, the recent focus of remote sensing applications has been to classify urban and built-up areas. Qiao et al.~\cite{5969075} demonstrate the use of classical techniques for object classification, such as a Support Vector Machine (SVM), to classify components of an urban area utilising high-resolution remote sensing imagery. Kanade et al.~\cite{9358951} demonstrate rapid mapping of urban regions for a high-density metropolitan city with varying built-up patterns, using remotely sensed data gathered from a synthetic aperture radar system. Li et al.~\cite{6946846} utilise aerial Light Detection and Ranging (LIDAR) with charge-coupled device (CCD) imagery to demonstrate an analytic hierarchy process for land area coverage classification, with the ability to differentiate between Urban and Non-Urban areas. Djamel et al.~\cite{9105142-2} evaluate classification schemes for urban area extraction using Landsat imagery and demonstrated that deep learning techniques outperformed traditional techniques for classification. Wen-mei Li et al.~\cite{8660387} compare a range of deep learning techniques to solve the environment-type classification problem in urban and built-up areas utilising remote sensing imagery. Man Li et al.~\cite{6394742} introduce hierarchical structuring to the land coverage classification problem to differentiate between land and water coverage from remote sensing imagery.

\subsection{Place Recognition}\label{sec:place_rec}
Place recognition is the ability \enquote{to recognise the exact place
despite significant changes in appearance and viewpoint}~\cite{VPR_where} within a map. Place recognition is commonly used throughout the literature to enable feature localisation within an occupancy grid ~\cite{DBLP:journals/corr/abs-2106-10458}. Barros et al.~\cite{DBLP:journals/corr/abs-2106-10458} compare deep learning approaches for place recognition, utilising methods on the spectrum of supervised to unsupervised learning categories. This survey also explores end-to-end frameworks used to address a domain translation problem for place recognition, for which NetVlad~\cite{NetVlad} offer the most considerable impact. The survey presents three approaches to supervised learning, consisting of holistic, landmark and region based. From this, landmark-based supervised training effectively solves appearance change, perceptual aliasing and viewpoint changing. The techniques demonstrated by Sunderhauf et al.~\cite{covnet_landmarks} employ convolutional neural network (CNN) feature extractors to develop semantic-based feature labels used for identifying potential landmarks in a visual feed. Rosinol et al.~\cite{Kimera_V2}, and Lajoie et al.~\cite{DOOR_Slam} utilise these extracted landmarks to generate a pose graph of key features within an environment.

\subsection{Map Matching}\label{sec:map_matching}
Map matching techniques are required in SLAM algorithms, enabling updates to prior environment models. As Williams et al.~\cite{WILLIAMS20091188} described, there are traditionally three main approaches: map-to-map, image-to-image and image-to-map. Integrating across a Gaussian distribution representation of a 6D voxels of an RGB-D sensor or point cloud system is one method to achieve spatial matching. Shuien et al.~\cite{map_merge_review} detail a networked solution for feature merging through map alignment and data association for applications with SLAM solutions on a MAS. Yufeng et al.~\cite{semantic_map_matching_2022} present a semantic labelling system integrated with spatial map matching to generate a more effective SLAM result for loop closure. MurArtal et al.~\cite{OrbSlam2} demonstrate SLAM techniques that utilised critical frame solutions that allowed map matching to occur as a data insertion into a graph rather than the spatial matching techniques previously used. The architecture developed by Andersone~\cite{State_of_the_art} extended this, allowing for cross-platform map merging between platforms that did not share the same sub-mapping techniques through semantic representation for common understanding. Kong et al.~\cite{8653820} demonstrate a ConvNet landmark-based visual place recognition system that utilises sequence search and hashing-based landmark indexing, which significantly increases the efficiency of the map-matching process. The place recognition architecture designed by Garg et al.~\cite{doi:10.1177/0278364919839761}conducts map matching through feature comparison by employing a semantic structure to generate an understanding of features. This system used CNN-based key-point matching, utilising semantic filtering and dense descriptor weighting to allow a place search procedure leading to a candidate match selection function.

\subsection{Ontology and Symbolic Reasoning}\label{sec:ontology}
Symbolic AI is the term used for several related AI methods that reason about problems using high-level human-understandable representations (symbols)~\cite{jean-baptiste_2021}. As Jean-Baptiste~\cite{jean-baptiste_2021} describes, an ontology is a \enquote{set of entities, which can be classes, properties, or individuals}, representing \enquote{complex knowledge sets about things and their relations}. One use of an ontology is to standardise the knowledge base of a specific domain and allow readability by humans and machines, as has been demonstrated by Baxter et al.~\cite{swarm_teaming_ontology}. Gomez-Perez~\cite{onto_engineering_2} describes that the standard components of an ontology ($\mathcal{O}$) are its instances ($I$), concepts ($C$), attributes ($a$), and the relationships ($R$) between them. Axioms ($A$) are developed within the ontology to generate assertions to describe the overall theory of the ontology for its application, thereby incorporating domain-level knowledge within the ontology. An ontology can be defined as the five-tuple~\cite{Onto4MAT}:

\begin{equation}
    \mathcal{O} = <C,R,a,I,A>.     
\end{equation}

An application of an ontology explored by Tenorth et al.~\cite{RoboEarth_4} is to standardise the semantic representation of language services available to cognitive agents within a system and create an ecosystem where knowledge between agents and operators is shared. This semantic representation of objects within the symbolic domain is essential for a shared language between individuals, required for communication to exist within a system~\cite{swarm_teaming_ontology}. Furthermore, enabling this communications layer within a system furthers the system’s capabilities to explain agent actions and hence provide explainability to an operator.
Hepworth et al.~\cite{swarm_ontology} explore the concept of system transparency and operator understanding throughout the proposed \emph{Human-Swarm-Teaming Transparency and Trust Architecture},  asserting that system interpretability and explainability are key facets underpinning the ability of a system to provide transparency for agent actions. Utilising these ontologies provides a method to share semantic knowledge, promoting bidirectional transparency to cognitive agents, be they artificial or human~\cite{Onto4MAT}. This transparency provides situational awareness of an autonomous agent’s actions, decisions, behaviours, and intentions to other agents within the system~\cite{chen2018}. Furthermore, as an ontology provides transparency and explainability within a system, it may enable agents’ joint function within MAS to complete a central role~\cite{doi:10.1177/0018720819879273}. Hence, an ontology may provide a semantic understanding of the world to the system, enabling a communication layer with cognitive agents through shared language services, thereby providing system transparency. 

Many fields have applied ontologies to incorporate expert-level domain knowledge into symbolic solutions. Utilising these hierarchical ontologies for guided learning in sub-symbolic systems is a concept explored by Campbell~\cite{campbell}. This application enables a sub-symbolic system to reason on abstract concepts and reduce the dimensionality of a problem space (through partitioning) by applying prior knowledge to a learning system; see, for example, Hepworth~\cite{Hepworth2021:ARS}, where a hybrid approach to activity recognition is detailed that fuses both data- and knowledge-driven (ontological) approaches to the activity recognition problem space in the machine learning domain. The RoboEarth framework~\cite{RoboEarth_3, RoboEarth_4} proposed a system that inherently integrates a knowledge base with visual SLAM, allowing a more accurate representation of the environment and recognition of the objects found within. The Triplet Ontological Semantic Model (TOSM)~\cite{TOSM} utilised short and long-term memory with a static ontology to create an on-demand ontological knowledge graph representing an environment in the symbolic domain. The abstract map data structure developed by Talbot et al.~\cite{9091567} incorporated symbolic spatial information (such as signs) into the SLAM problem to create \enquote{malleable spatial models for unseen places}, producing navigation results comparable to the ability of humans. Control agents were utilised within the HST-3 architecture~\cite{swarm_ontology} to mitigate against heterogeneous knowledge sets produced by a swarm, which hinders convergence to a central solution. The current, most advanced ontology designed specifically for SLAM tasks developed by Cornejo-Lupa et al.~\cite{robotics10040125} combines data from a range of ontology sets to achieve superiority at the domain knowledge, lexical and structural levels. The onto4MAT is the first attempt to design an ontology for multi-agent teaming~\cite{Onto4MAT} systematically. It enables an operator to provide an intent as tasks to a multi-agent system and for the agents to provide feedback to the operator.

\subsection{Critical Assessment of Symbolic Approaches for Environment Type Classification}\label{sec:SymboSLAM}
The present literature on environment-type classification relies heavily on remote sensing techniques to provide a high-fidelity overview of an area. Current technology in this field utilises techniques similar to those found within the object detection realm to classify environments through pattern recognition from these area overviews. However, little to no literature on symbolic approaches for environment-type classification. Furthermore, there is little to no literature on utilising SLAM techniques for environment-type classification. The SymboSLAM architecture introduces a method that combines the feature extraction strengths available to sub-symbolic AI architectures with symbolic approaches for reasoning to develop a system capable of conducting environment-type classification, thereby bridging this gap within the literature.

\section {Methodology}\label{sec:methodology}

\subsection{Project Overview and Scope} \label{sec:scope}
The SymboSLAM architecture is a hybrid edge-driven context reasoning approach for use on a MAS of intelligent edge agents. This architecture produces environment-type classifications for an area, presented as 2D ground maps with features and landmarks represented symbolically. Environment-type classification is made possible by transforming the environment into a set of state-space variables that are semantically queryable. Subtasks 1-3 below show the general scheme for this transformation:

\begin{enumerate}[label=T\arabic*]
    \item \textbf{Feature extraction.} This sub-task enables the system to extract useful information from the environment and generate an information pipeline. The edge agents on the MAS architecture achieve this through feature extraction and localisation.
    \item \textbf{State Space Maps.} This task takes the information pipeline from T1 and generates useful semantically labelled maps. Both the edge agents at an individual level and the control agent at a collective level conduct this subtask. This task is deployed on a MAS and is achieved through map-matching / map-merging techniques.
    \item \textbf{Ontology.} This sub-task takes the semantically labelled state space representations of the environment developed from T2 and formulates the labels for segments of the presently mapped environment. The control agent achieves this task through an ontology to achieve symbolic reasoning and hence enable transparency in the meaningfully labelled maps of the environment.
\end{enumerate}

We propose a hierarchical structure with edge agents reasoning about their environment, employing sensor-based and visual data sources. The proposed architecture utilises a blend of sub-symbolic and symbolic reasoning techniques. First, sub-symbolic AI modules are used for object detection to achieve feature extraction. Next, the extracted features are semantically labelled and placed within a pose graph, transforming the environment into a set of queryable state space variables. Finally, a symbolic AI module for context reasoning then works to infer the environment type of segmented portions of a map on a collectively referenced coordinate system. The symbolic component queried to determine the environment type is an ontology.

\subsection{Edge Agent Search Method}\label{sec:arc_des}
\begin{figure*}[ht]
    \centering
    \includegraphics[width=\textwidth]{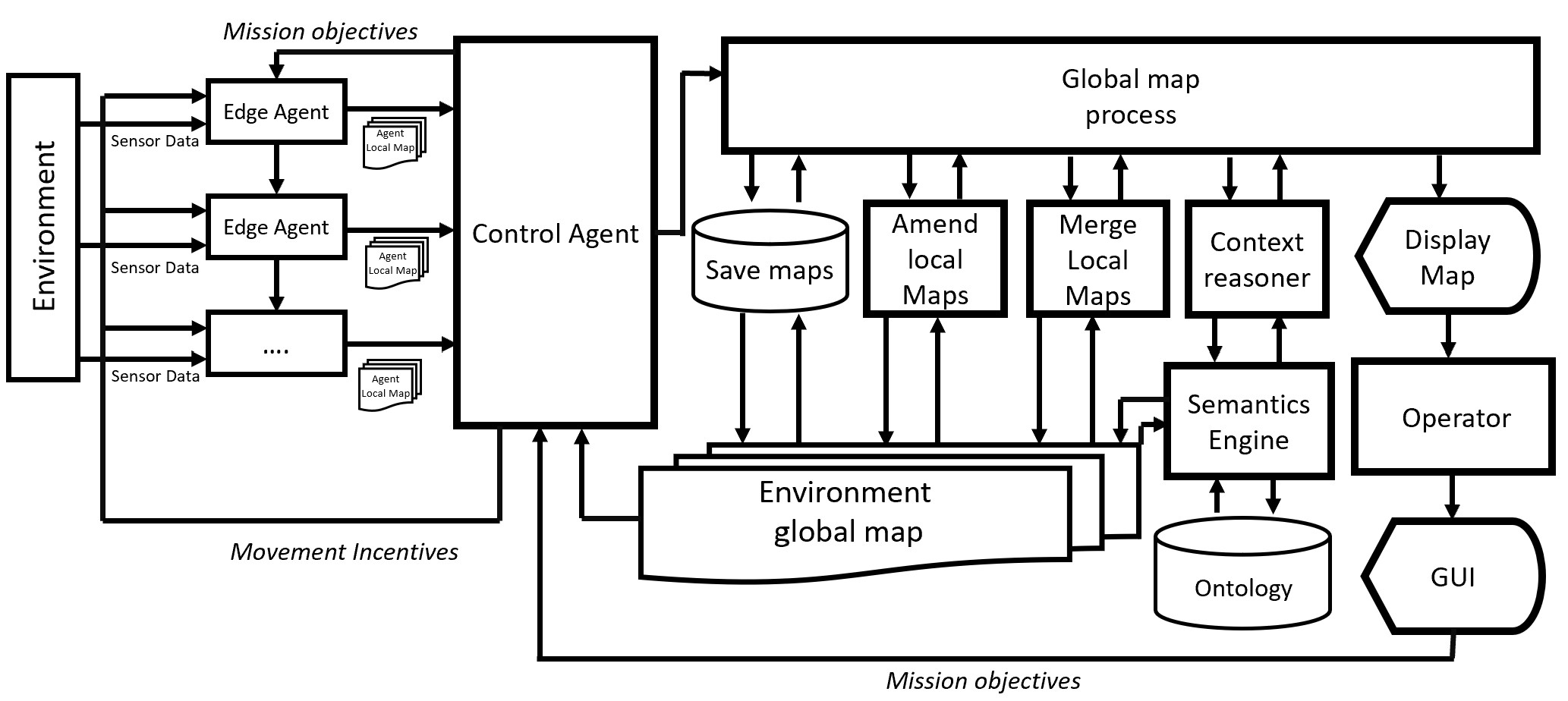}
    \caption{SymboSLAM control agent architecture.}
    \label{fig:control_agent}
\end{figure*}

The simulated implementation utilises a random walk search strategy that extends the random-tree search described by Washburn~\cite{washburn_2014} to introduce reactionary and long-range coordinated targeting strategies through movement incentives to achieve a more significant distribution level for higher area coverage. The proposed hierarchical control agent architecture shown in Figure~\ref{fig:control_agent} will coordinate the swarming movement of the edge agents through movement incentives similar to the shepherding behaviour presented by Hepworth et al.~\cite{footprints} to effectively disperse the search agents throughout an environment for target search and SLAM. A human operator was utilised as the cognitive agent for search methodology to instantiate the edge agent architecture on a physical device.

\subsection{Feature Extraction} \label{sec:feat_extract}
The CNN feature extraction submodule detects 56-class objects selected to model the Canberra landscape across the seven possible environment types available. Feature abundance in the environment and access to objects in existing datasets were the determining factors for selection into the SymboSLAM custom dataset. Image-classification pairs were taken from 10 of the most popular datasets available. These were combined with the SymboSLAM dataset to generate 200k, 12k, and 6k labelled images for training, validation and testing.

\subsection{Place Recognition} \label{sec:place_recognition}
A landmark-based technique is employed at both the edge agent and control agent architecture level to achieve the place recognition component of the SLAM problem whilst building a queryable state space representation of the environment. Ground vehicles controlled by the control agent architecture explore within an environment employing the random walk search strategy described above, conducting locally-scaled SLAM tasks. Each edge agent constructs an individual map of its environment, utilising onboard Spatio-temporal sensor feeds. The simulated ground vehicles utilise a camera, Inertial Measurement Unit (IMU), Time of Flight (ToF) and Lidar sensors, sequentially timestamped; the implementation on the android application uses a camera and GPS. Each edge agent implements a feature extraction submodule described above. The edge agent architecture then utilises semantic knowledge of these extracted features to sub-categorise objects into either dynamic or static objects. Landmarks within sub-maps used for matching represent the collected static features, and both the static and dynamic objects are utilised in later recognitions for environment-type classification. The depth sensor is then utilised for the simulated ground vehicles to determine the closest of the static targets, and the edge agent will then navigate towards it. The edge agent continues on its path towards this target, attempting to minimise the proximity to the target whilst maintaining the target within the entire frame of the camera feed, as demonstrated in algorithm~\ref{alg:alg_target_acquisition}. Once the distance between the agent and the target is minimal, the edge agent utilises its onboard depth sensor to determine the feature's location relative to the agent's location. Finally, the feature is inserted into the edge agent's individual map data structure as an element within a pose graph.

 \begin{figure*}[ht]
    \centering
    \includegraphics[width=\textwidth]{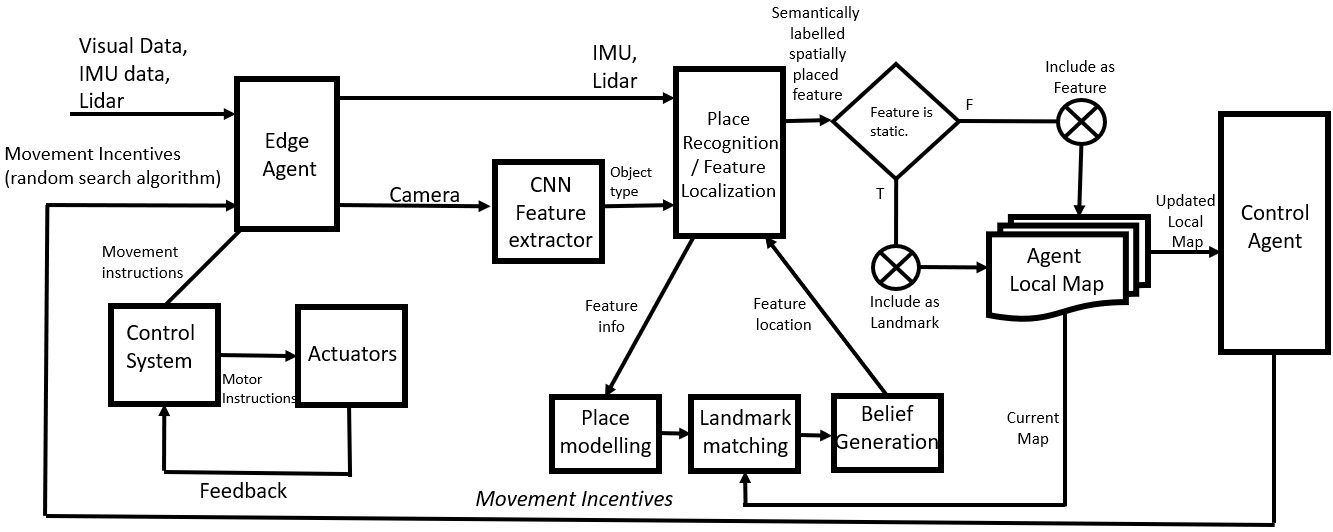}
    \caption{SymboSLAM Edge Agent architecture for a simulated robotic platform. Note that the physical instantiation of an edge agent holds the same basic topology but utilises GPS instead of an IMU.}
    \label{fig:edge_agent}
\end{figure*}

\subsection{Map Matching} \label{sec:map_matching_method}
\begin{figure*}[ht]
    \centering
    \includegraphics[width=\textwidth]{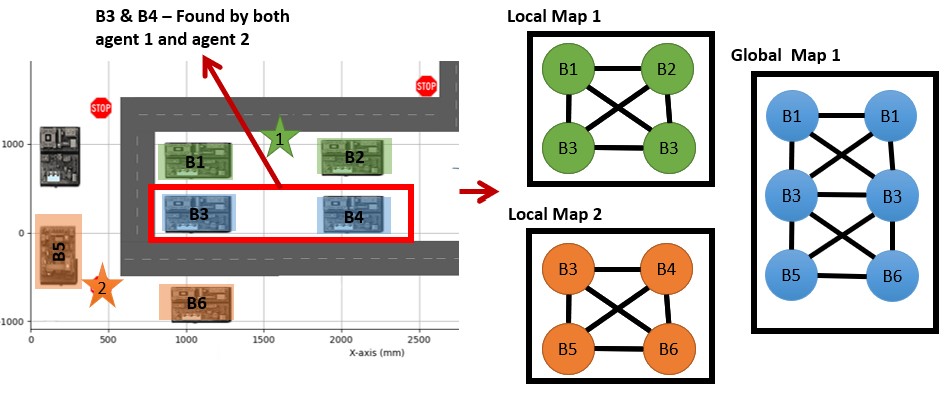}
    \caption{Map matching technique employed by the SymboSLAM architecture at both an individual and collective level. Note that the nodes shown within each map are an entry within a pose graph and then displayed to the operator as a point on a 2D occupancy map.}
    \label{fig:map_match}
\end{figure*}

A map-to-map, map-matching technique utilising landmarks as key features within a pose-graph data structure is the technique featured in both edge and control agent architectures for individual and collective mapping. Figure~\ref{fig:map_match} demonstrates this functionality at the control agent level to conduct map meshing. The control agent receives and maintains the individual maps for each edge agent. Each of these individual map data structures is referenced to begin at $(x,y) = (0,0)$. The control agent then amends the received individual maps to reference them on the collective map referencing plane, as the control agent knows the true starting location of each edge agent. The edge agents' map features are then merged onto the control agents' map utilising the landmark description and believed place. The control agent compares the landmark descriptor of each new landmark received with all features within some radius $a$ within the control agent's map to determine whether they are semantically similar. If they are semantically similar, the feature with the highest confidence is the new landmark, and this landmark position is the average of the two previous landmark positions. Suppose the two landmarks are not semantically similar, depending on the tolerance of feature proximity and the feature confidence. Then, the new landmark will either be added to the pose graph as a new entry or discarded. The discarded landmarks are stored, and if there are greater than two discarded landmarks within close spatial proximity (tolerance again specified by operator) with similar classes, then the landmark at that position will be updated with the discarded feature.

\subsection{SymboSLAM Ontology} \label{sec:SymboSLAM_Ontology}
\begin{figure*}[ht]
    \centering
    \includegraphics[width=0.75\textwidth]{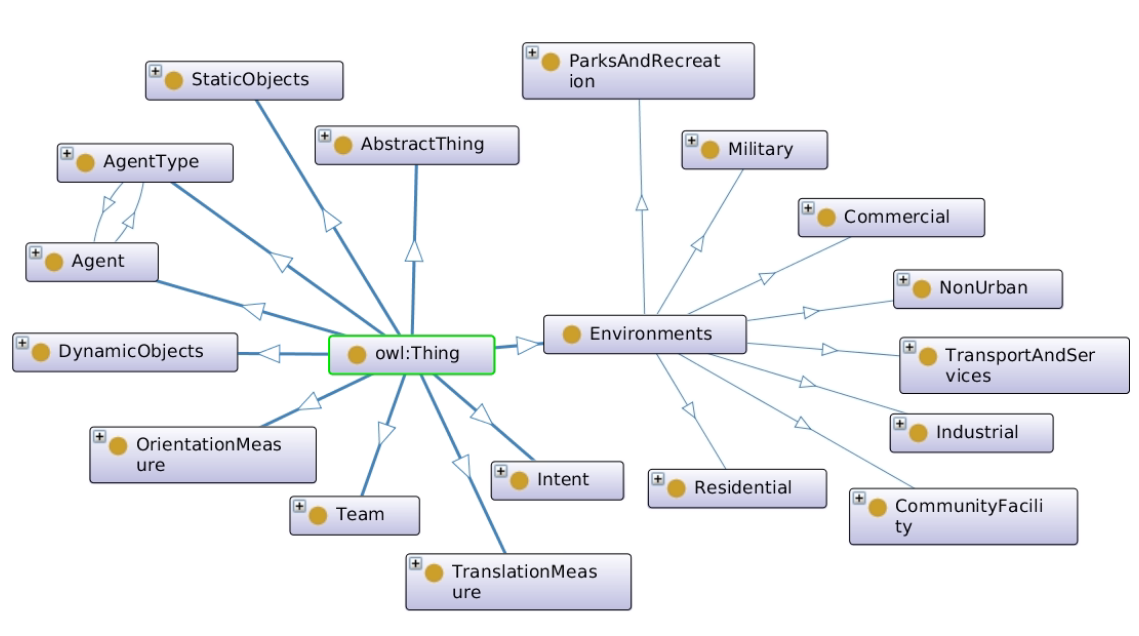}
    \caption{SymboSLAM ontology. Environment types entries reflect the Canberra region. The SymboSLAM ontology also features concepts from the OntoSLAM and ONTO4MAT ontologies to enable more effective SLAM and swarm control functionality, respectively.}
    \label{fig:onto1}
\end{figure*}

The SymboSLAM ontology, depicted in Figure~\ref{fig:onto1}, is designed for the SymboSLAM architecture. This ontology has domain-level knowledge for environment contextualisation and subsets of the Onto4MAT~\cite{Onto4MAT} and OntoSLAM~\cite{robotics10040125} ontologies for swarm control and SLAM problems. This ontology enables bidirectional communication between all cognitive agents (edge, control, operator) by allowing the system to conduct SLAM tasks with symbolic representations. A single control agent is used to ensure that the explainability of a solution is converged upon, enabling transparency in the developed system to foster trust by the operator.

\subsection{Semantics Engine} \label{sec:semantics_engine}
\begin{figure}[ht]
    \centering
    \includegraphics[width=\textwidth/2]{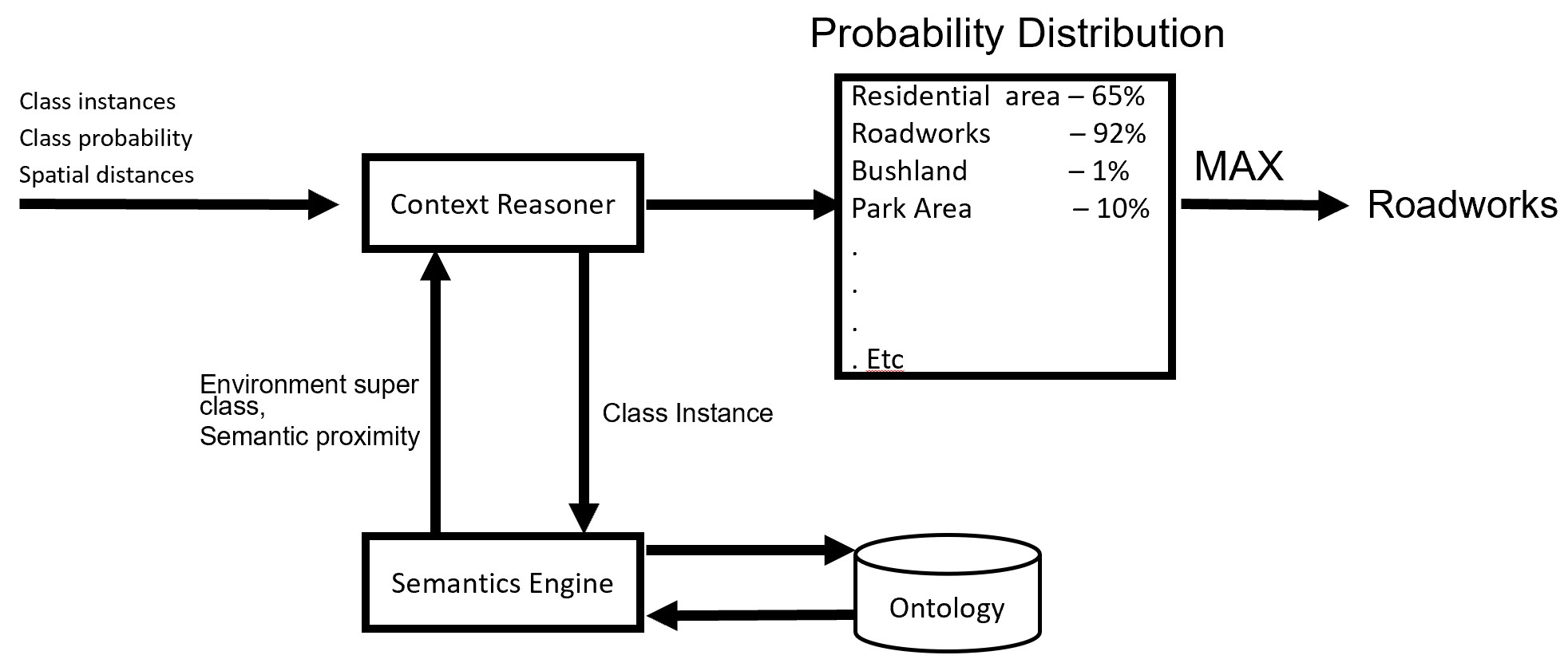}
    \caption{SymboSLAM symbolic reasoning modules. The ontology featured is the SymboSLAM ontology.}
    \label{fig:sem_eng}
\end{figure}

Figure~\ref{fig:sem_eng} shows the submodule responsible for environment classification through the contextualisation of the features found within a segment. For each segment of the partitioned map, this module takes the feature class instance, feature class confidence and spatial distances between each feature over a segment. It outputs a probability distribution of environment types for which the maximum probability within this distribution is this segment's environment type. The SymboSLAM ontology is queried with a feature class instance to determine the environment type superclass and the semantic proximity to each environment superclass; note that multiple environment types may be returned. This module intends to reward:
\begin{enumerate}
    \item semantic closeness of feature class and environment superclass; 
    \item spatial closeness of feature classes with alike environment superclasses; and 
    \item a larger number of inferences made between feature classes within a segment, for instance, more feature classes identified with alike environment superclasses. 
\end{enumerate}

The semantics engine module penalises the spatial closeness of feature classes with dissimilar environment superclasses. For each feature in a map segment, the semantic proximity for each possible feature environment super class is $SP(e_n, f_x)$. The distances between each feature class within the segment is then calculated to be $d(f_x,f_y)$ and each distance is normalised to the maximum possible distance of each segment such that $d(f_x,f_y) \sim N(0,1)$. The confidence that each segment is of environment n is calculated by summing the normalised distance between each feature node of that environment type weighted by the confidence of each feature class:

\begin{equation}
    C(e_n) = \sum_x \sum_y (SP(e_n, f_x)\cdot f_x +  SP(e_n, f_y)\cdot f_y  ) \cdot (1-d(f_x,f_y))
\end{equation}

The environment confidence is calculated for each environment $n$ contained within the SymboSLAM ontology to produce a probability distribution across all possible environment types for each segment. The confidence of environment type is then normalised to the number of inferences made between the features $f_x, f_y$ for each environment confidence calculation so that $C(e_n) \sim N(0,1)$. Equation 2 above solves reward intents one and two listed above. The semantics engine then calculates a weighted sum for the environment confidence $C(e_n)$ and the number of inferences made for each environment confidence calculation normalised to the maximum number of inferences allowable for a set of segment features. This calculation solves reward intent three and produces a probability distribution of the likeliness of environment type. For example, for a segment with $z$ features, this is given by:

\begin{equation}
 P(e_n) =\frac{1}{a} \cdot C(e_n) + \frac{1}{1-a} \cdot \frac{e_n(\text{inferences})}{z(z-1)} 
\end{equation}

Where $a, C(e_n), P(e_n) \sim N(0,1)$ and $a$ are specified by the operator depending on the sparsity of feature classes within a map, as some environment types may inherently have a low number of features, e.g. a desert environment. 

\subsection{Grid Map Segmentation}\label{sec:map_seg}
\begin{figure*}[ht]
    \centering
    \includegraphics[width=0.9\textwidth]{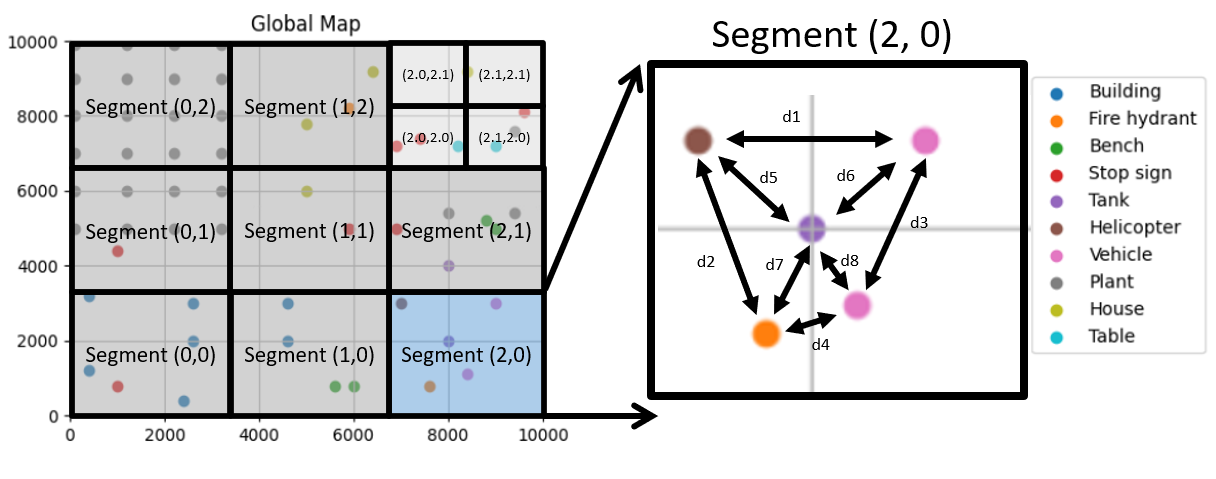}
    \caption{Segmentation process for map partitioning - Segment (2,2) shows further partitioning where the confidence of environment classification is less than a determined threshold}.
    \label{fig:seg_map}
\end{figure*}

The control agent attempts to increase the environment-type classification probability through map segmentation methods. Figure~\ref{fig:seg_map} depicts the functionality of this module (left) and shows the output of the segmentation process (right), Which the semantics engine module then utilises. The control agent first identifies nine critical regions of interest (ROI) within the map. The semantics engine then conducts environment-type classification using this updated segmented map data structure. If the probability distribution for environment type classification does not return above a specified threshold, then the segment is further partitioned into four quadrants, as shown in Segment-$(2,2)$ of Figure~\ref{fig:seg_map}. This process is repeated until one of three conditions are met. The first is if the quadrant of a segment returns an environment classification confidence above a specified threshold, and the next is if the quadrant is empty. The third is if the sparsity of information within the quadrant reduces classification performance. 

\subsection{N-Nearest Neighbour Map Segmentation}\label{sec:NNN-seg}
The second segmentation methodology implemented utilises the environment probability calculation from the semantics engine. This segmentation methodology is based on the $K$-nearest-neighbour algorithm and is a novel progressive clustering algorithm. The user supplies this algorithm with a seeding position from the control agent's map (where to begin clustering) and a measure of momentum (how many landmarks the algorithm skips between clustering), from which the algorithm will conduct parses of the control agent map. The branch segmentation method then uses the three landmarks closest to the cluster centre to determine its environment probability distribution. The number of landmarks within a cluster is incremented by one until the prior environment probability stored is larger than the posterior environment probability.

The $N$-Nearest neighbour algorithm aims to cluster the entire landmark map into $N$ previously unknown fragments, each with a unique number of landmarks. The branch segmentation algorithm then compares the neighbours of these $N$ fragments to merge neighbours with the same environment classifications. The above algorithm essentially takes some number of Landmarks $L$ within a Map $M$:

\begin{equation}
     M=\{L_0, L_1, ... L_\text{end}\} \mbox{  , where  } L\in 	\mathbb{R} ^D
\end{equation}

And places them into a fragmented map set:
\begin{equation}
     M=\{F_0, F_1, ... F_z\}
\end{equation}
$ \mbox{Where } F=\{L_0, L_1, ... L_N\} \mbox{ and } F_\text{env} = \text{argmax}(P(e_n))$

To ensure fragmented sections of the map did not bisect, the control agent used a modified SVM algorithm to create decision boundaries surrounding clusters. As map fragments may have complex shapes, landmarks were compared as pairs to check for bisects with landmarks from neighbouring fragments. The modified SVM algorithm used the landmarks immediately surrounding the bisection portion of neighbouring fragments to draw the decision boundary. The control agent then coordinated trade between neighbouring clusters based on these decision boundaries. 

\begin{equation}
    \mbox{ For } F_a \xrightarrow[]{} L_{a_1},  L_{a_2} \mbox{  and  } F_b \xrightarrow[]{} L_{b_1}, L_{b_2}     
\end{equation}
\begin{equation}
    L_{a_1}, L_{a_2} \xrightarrow[]{} \text{line}_1,  \mbox{ and } L_{b_1}, L_{b_2} \xrightarrow[]{} \text{line}_2 
\end{equation}

Then given line$_1$ and line$_2$ bisect, a decision boundary can be drawn between two sets of landmarks from fragments A and B:

\begin{equation}
    L\in \mathbb{R} ^D \mbox{ , } \phi: \mathbb{R} ^D \xrightarrow[]{} \mathbb{R}^M
\end{equation}

\begin{equation}
    \phi (L) \in \mathbb{R}^M
\end{equation}

\begin{equation}
    H : w^T \phi (L) +b =0
\end{equation}

And the distance between the decision boundary and the landmarks within a segment's immediate neighbour cluster (excluding the bisecting points) is found with:

\begin{equation}
    d_H(\phi(L_0)) = \frac{|w^T \phi(L_0)+b|}{||W||_2}
\end{equation}
 
The decision boundary is then updated to maximise the minimum distance from the decision boundary hyperplane to each landmark point in a map segment according to the following:

\begin{equation}
    W^*=\underset{\text{w}}{\text{argmax}}\frac{1}{||W||_2}[\underset{\text{q}}{\text{min}}\cdot y_n \cdot [W^T \phi (L_q) +b]]
\end{equation}

\subsection{Full SymboSLAM architecture}\label{sec:NNN-seg}
The SymboSLAM architecture synthesises practices from the SLAM, swarm and symbolic domains to conduct environment-type classification in a novel manner. The SymboSLAM architecture transforms the environment into a query-able state space representation through a Multi-Agent System featuring a hybrid reasoning orientation. The SymboSLAM architecture utilises intelligent edge agents capable of processing information to collect data and infer knowledge about their environment. These edge agents apply ontologically backed semantic labels to extracted features, thereby generating symbolic representations of their environment. Sub-symbolic methods for feature extraction are used in conjunction with place recognition techniques to generate area overview maps with semantic markers embedded within the spatial information provided. Map matching techniques are then incorporated within the architecture to merge these semantically labelled individual maps produced by intelligent edge agents into a central structure. A control agent collates these individual maps and generates a collective map to infer knowledge about the environment. An ontology is then featured within a symbolic reasoning approach to take the semantically labelled representations of the features within this central map structure and produce a 2D map of environment-type classifications.

\section {Architecture Evaluation}\label{sec:Eval}

Nine metrics were selected to evaluate the SymboSLAM architecture, as shown in Table~\ref{fig:eval_metrics}. These evaluation metrics reflect the specific domain from which each submodule was derived. The evaluation metrics are described below. The studies shown in Table~\ref{fig:eval_metrics} reference instances from work published within the same domain as the submodule being evaluated for which these metrics have previously been used.  

\begin{table*}[t]
    \centering
    \resizebox{\textwidth}{!}{
   
    \begin{tabular}{ |p{10cm}|p{3cm}|}
     \hline
     \multicolumn{2}{|c|}{Architecture Evaluation metrics - Studies and their respective metrics } \\
     
     \hline
     Type of metrics & Studies \\
     
     \hline 
     
     \textbf{Multi Agent System} & \\
     
     \hline
     
     Area Coverage - $\sfrac{A'}{A}$ scoring & \cite{washburn_2014, robot_swarms_search, Chen2020ActiveSF}\\
     Area Dispersion - average dispersion measure & \cite{abbass_hunjet_2022, footprints}\\
     Time - average time to complete & \cite{Kimera_V2, magic, TOSM}\\
     
     \hline
     
     \textbf{Simultaneous Localisation and Mapping} & \\
     
     \hline
     
     Feature Extraction - mAP & \cite{yolov3tiny, Bochkovskiy2020YOLOv4OS, Yolo_Slam_2}\\
     Place Recognition - ground truth comparison  & \cite{Schwertfeger2013EvaluationOM, covnet_landmarks, DOOR_Slam, Kimera_V2}\\
     Map Matching - average error of centre offset & \cite{8653820, DBLP:journals/corr/abs-1804-05526, doi:10.1177/0278364919839761}\\
     
     \hline
     
    \textbf{Symbolics Engine} & \\
     
     \hline 
     
     SymboSLAM Ontology - OOPS! & \cite{poveda2014oops, Onto4MAT, onto_engineering}\\
      Map Partitioning - IoU / Kappa Coefficient & \cite{6927922,  9105142-2, 8660387}\\
     Area Type Classification - AP & \cite{8071701, 6927922, 5969075, 9358951, 8660387, 6394742}\\

     \hline
    \end{tabular}
  
    }
\captionof{table}[Metrics]{Nine metrics utilised to evaluate the SymboSLAM architecture with published studies that have previously used these evaluation metrics}
\label{fig:eval_metrics}
\end{table*}

\subsection{Multi-Agent System}\label{sec:MAS_Eval}
The area coverage, area dispersion and time taken by the architecture to complete are the metrics used for evaluating the multi-agent system component of the SymboSLAM architecture. These evaluation metrics were tested in simulation only, as no robotic edge agent platforms were used for physical instantiations. $\sfrac{A'}{A}$ scoring as described by Washburn~\cite{washburn_2014} is the evaluation metric for area coverage, where $ A' =VWt $ is the searched area of the maximum possible searchable area A. The Agent dispersion from the global centre of mass, as described by Abbass et al.~\cite{abbass_hunjet_2022}, was taken as the success metric area dispersion. The dispersion was measured at intervals of 1 second throughout the simulated trials, and an average of this data was taken to determine agent area dispersion. The simulated trials concluded when all objects were discovered by an edge agent within the environments presented. The measure of area coverage, area dispersion and time taken to complete was calculated after the simulated trial.

\subsection{Simultaneous Localisation and Mapping}\label{sec:SLAM}
As the feature extraction submodule consists of a CNN object detection system, the intersection over union (IoU) and the mean average position (mAP) for the feature extraction submodule were tested as this unit's first evaluation metric. The IoU for each bounding box for a detected object class in a frame is:

\begin{equation}
    IoU = \frac{\mbox{area of overlap}}{\mbox{area of union} } = \frac{X \cap Y}{X \cup Y}
\end{equation}

Where $X$ is the detected feature bounding box, and $Y$ is the ground truth. The precision and recall of the model to detect objects are given by

\begin{equation}
    \text{Precision} = \frac{TP}{TP + FP} \mbox{ , } \text{Recall} = \frac{TP}{TP + FN},
\end{equation}

$\mbox{where TP \& FP = true \& false positive, FN = false negative}$

The average precision is determined by finding the area under a precision-recall curve when conducted over several prediction outputs: 

\begin{equation}
    AP = \int_{0}^{1} p(r) \,dr
\end{equation}

And lastly, the mAP is found by taking the mean of the AP over all the feature classes for which the model is trained.

\begin{equation}
    mAP = \frac{AP_1, AP_2, ... AP_n}{n}
\end{equation}

The SymboSLAM feature extraction module is evaluated against the testing dataset of 6k images. The place recognition component of the SLAM problem is assessed through ground truth map comparison, where maps are evaluated using matched topology graphs. As the ground truth and generated maps are already 2d topology graphs, their evaluation in the method described above can be directly applied. Next, the coverage (percentage of matching vertices between ground truth and generated map) and global accuracy (Correctness of positions of the matched vertices in the collective reference frame) were determined and used as the metric for place recognition. Lastly, the average error of centre offset is the evaluation metric used for spatial consistency of matched landmark pairs. The average error of centre offset is  a sum of squared error measure for the distance between the believed location and the actual location of all landmarks within a generated occupancy map:

\begin{equation}
    Er = \frac{1}{L}\sum_{i}^{L-1} \sqrt{(L_{ix} - L_{(i+1)x})^2 + (L_{iy} - L_{(i+1)y})^2}
\end{equation}

Where L is a landmark within a map.  

\subsection{Symbolic Engine}\label{sec:symbolic_engine}
The SymboSLAM ontology was evaluated using OOPS!~\cite{poveda2014oops}\footnote{https://oops.linkeddata.es/response.jsp} as demonstrated by Hepworth et al.~\cite{Onto4MAT}. This evaluation saw the checking of requirements and competency questions and testing of the ontology in the target application environment. The IoU and the AP between the generated 2D environment-type classification maps and the ground truth environment-type maps are calculated for all 16 simulated and measured trials. The IoU was used as the correlation coefficient (similar to a Kappa coefficient) to determine the inter-rater reliability between the generated and truth environment type map. The AP was used to determine the accuracy of the symbolic engine of the SymboSLAM architecture.

\subsection{Simulated Evaluation}\label{sec:Simulated_Evaluation}
There is \enquote{little prior work on symbolic navigation of unseen places and no relevant benchmarks for evaluating performance}~\cite{9091567} and less still on applying symbolic or sub-symbolic AI to the environment contextualisation problem. Therefore, simulated environments incorporated expert domain-level knowledge for classifying an environment type through contextualisation using the feature classes found within. The SymboSLAM architecture and ontology were deployed on the EyeSim robotics simulation software, employing three edge agents and one control agent to conduct environment contextualisation in real time. In addition, the SymboSLAM SLAM modules incorporated an oracle system to offset the errors from the simulated sensor feeds introduced by the Unity physics engine, causing a cascading effect due to the quasi-random dispersion SLAM algorithm. Six areas were simulated from the Canberra region to evaluate the SymboSLAM architecture within the simulated trials. These areas include Gunghalin (mix of environment types), the airport area (primarily non-urban), Fyshwick (industrial principally), Kingston (primarily commercial and high-density residential), the Train Depot area (mainly transport and services) and the Civic City precinct (commercial principally). Each simulation contained up to 56 separate 3D object classes, some classes with multiple models and all features with various instances.

\begin{figure}[ht]
\centering
    \includegraphics[width=\textwidth/2]{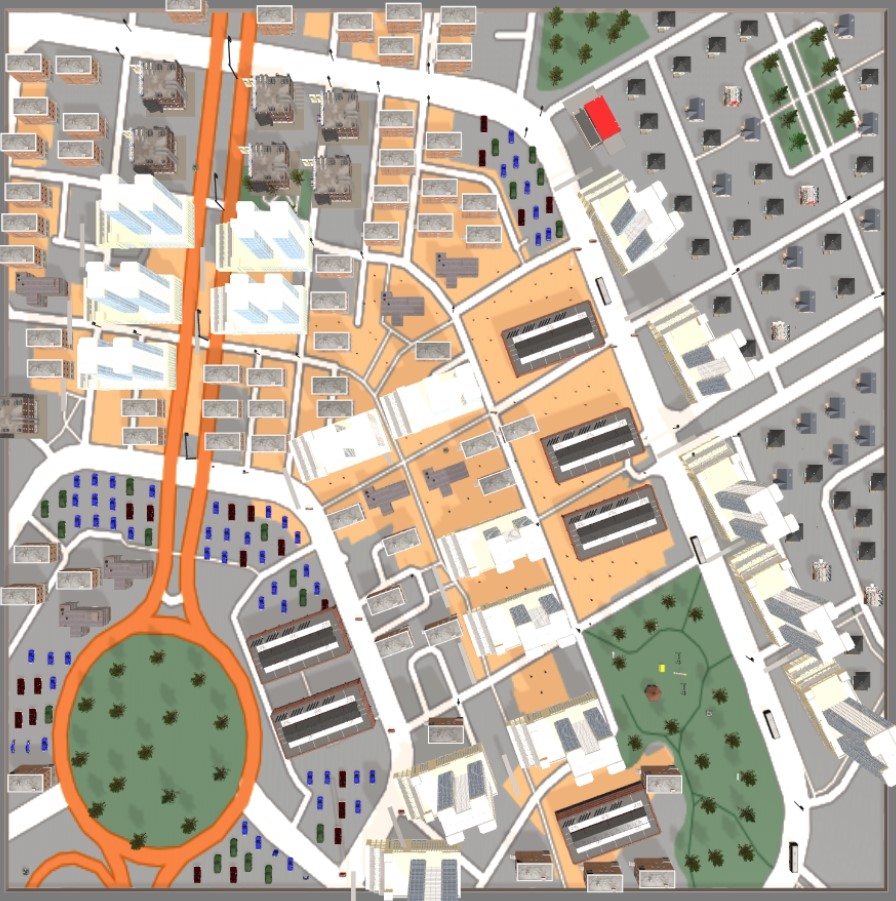}
    \caption{Example simulated area using the Unity 3D environment, depicting a region of Civic in Canberra.}
    \label{fig:civic}
\end{figure}

\subsection{Real World Experiments}\label{sec:Real_world_experiments}
Two real-world experiments were conducted using the SymboSLAM mobile phone application, acting as edge agents to collect landmark information. These experiments were conducted throughout the Civic and Gunghalin areas to showcase the functionality of the architecture in a real-world setting. 

\section {Results}\label{sec:Results}

\subsection{Multi-Agent System}\label{sec:feat_extract_results}
Table~\ref{fig:MAS_Results} shows the results for simulation area and area coverage, average and normal dispersion and time taken for the simulation to conclude. Note that the normal dispersion was calculated by taking the average dispersion of all edge agents from the GCM (column 5) and normalising it to the maximum theoretical dispersion to the GCM.


\begin{table*}[t]
\normalsize
    \centering
    \resizebox{\textwidth}{!}{
   
    \begin{tabular}{ |p{3cm}||p{3cm}|p{3cm}|p{3cm}|p{3cm}|p{3cm}|p{3cm}|}
     \hline
     \multicolumn{7}{|c|}{MAS Metric Results} \\
     \hline
    Simulation trial 
    & Total Area ($A$, km$^2$) 
    & area Searched ($A'$, km$^2$) 
    & Area Coverage($\sfrac{A'}{A}$)
    & Avg Dispersion (km)
    & Norm Dispersion 
    & Time (mm:ss)\\
     \hline
     01 - Gunghalin  & 29.5609 & 26.4414 & 89.44\% & 2.9945 & 77.88\% & 84:14  \\
     02 - Airport & 27.8891 & 20.7748 & 74.49\% & 2.6738 & 71.60\% & 43:39  \\
     03 - Fyshwick & 6.6667 & 5.5463 & 83.19\% & 0.8567 & 46.92
\% & 30:14  \\
     04 - Kingston & 0.5041 & 0.4423 & 87.74\% & 0.2432 & 48.44\% & 14:37  \\
     05 - Train Depot & 1.0976 & 0.9777 & 89.07\% &0.4452 & 60.10\% & 12:43  \\
     06 - City  & 1.8225 & 1.7943 & 98.45\% &  0.3098 & 32.45\% & 40:22  \\
     
     \hline
    \end{tabular}
    
    }
\captionof{table}[MAS_Results]{3 EyeSim robots were placed at equidistant points within each simulation listed. These edge agents collected feature information and were controlled by a control agent. The simulation concluded when all features from within the environment were discovered}
\label{fig:MAS_Results}
\end{table*}


\subsection{Simultaneous Localisation and Mapping}\label{sec:state_space_res}
\begin{figure}[ht]
\centering
    \includegraphics[width=\textwidth/2]{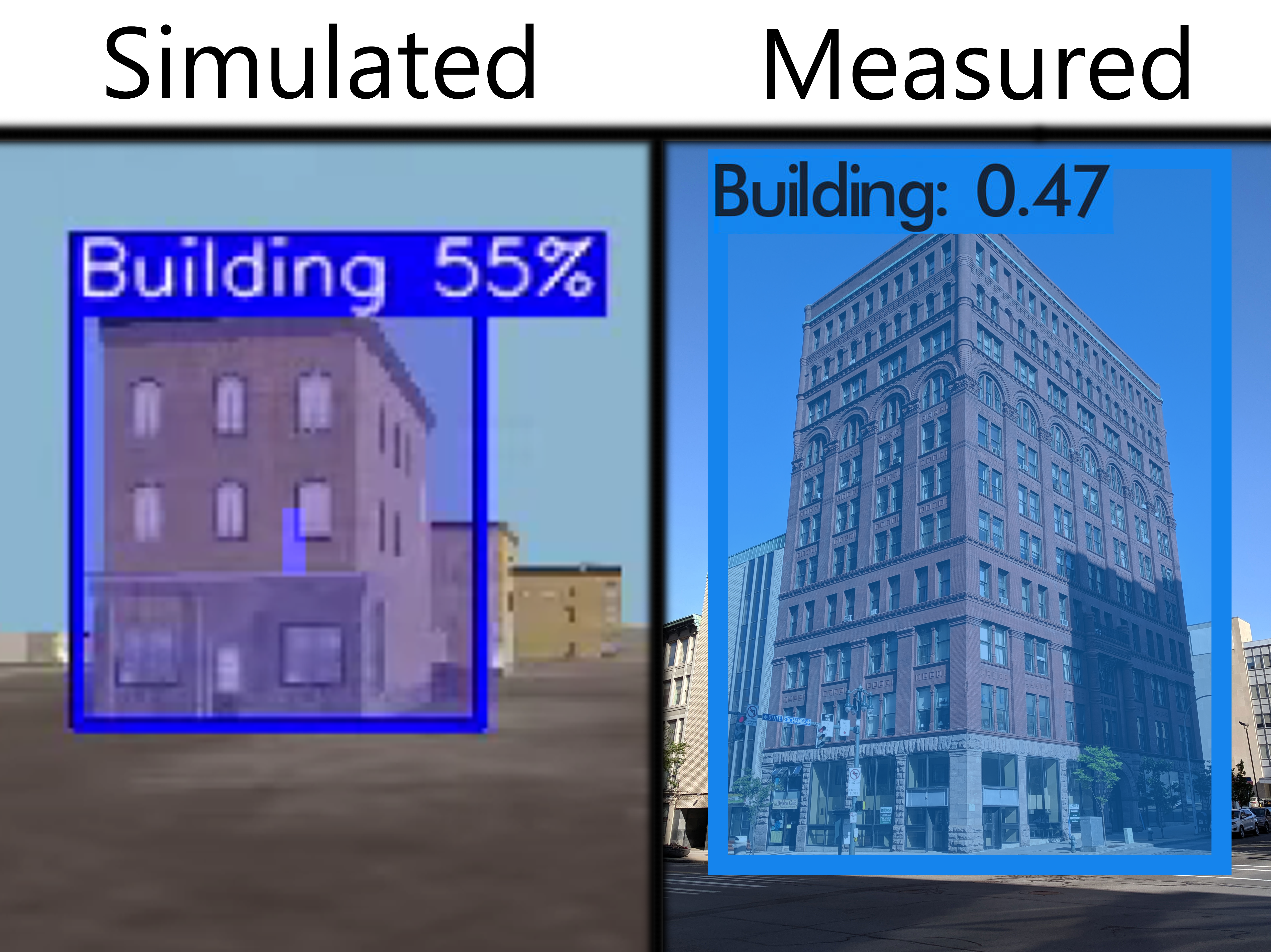}
    \caption{CNN Feature extractor predictions output from simulated (left) environment and real-world (right) environments}
    \label{fig:pred_output_sim}
\end{figure}

\begin{figure}[ht]
\centering
    \includegraphics[width=\textwidth/2]{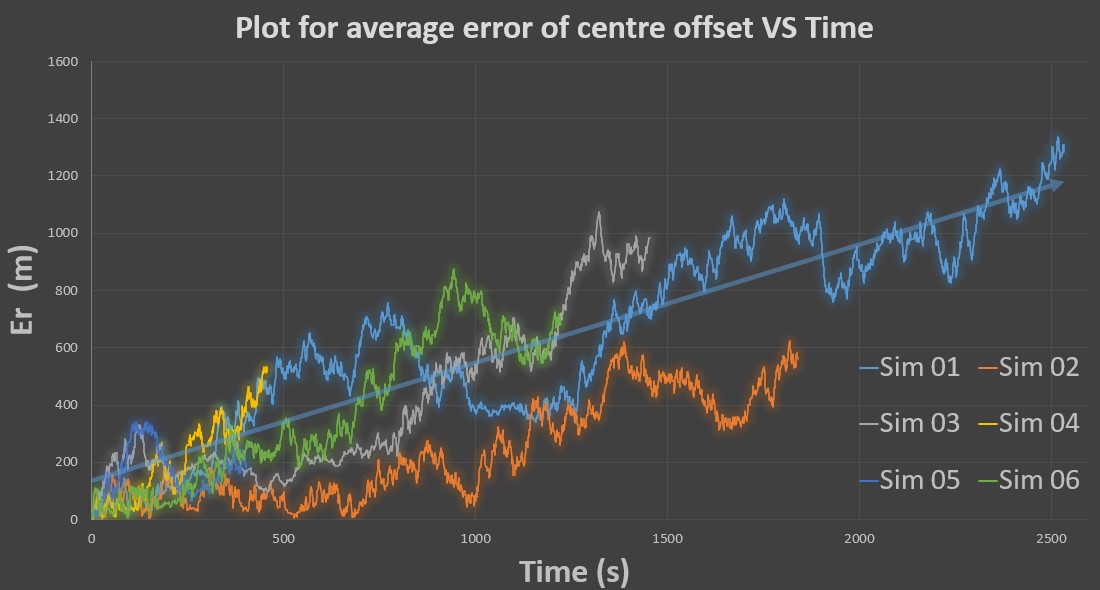}
    \caption{The average error of centre offset for Landmarks discovered by Edge agents when conducting map matching with ground truth maps for Simulations 1-6}
    \label{fig:avg_error}
\end{figure}

The 6k image testing dataset was utilised to evaluate the full-scale and tiny-scale CNN feature extraction module for the simulated and real-world SymboSLAM instantiations, respectively. The results for mAP, IoU and inference time are shown in Table~\ref{fig:CNN_Results}. A demonstration of the full-scale simulated instantiation is shown in Figure~\ref{fig:pred_output_sim} (left) and an example of the tiny-scale real-world instantiation is shown in Figure~\ref{fig:pred_output_sim} (right), and a full set of simulated and real-world results are available online~\footnote{https://cloudstor.aarnet.edu.au/plus/s/0T0pZFYgeyMgMyP}. Figure~\ref{fig:avg_error} shows the average error of centre offset for discovered landmarks. The exponential style in which this error accumulated led to the need to utilise an oracle system to test the functionality of the remaining SymboSLAM submodules. 

\begin{table}[t]
\normalsize
    \centering
    \resizebox{\textwidth/2}{!}{
   
    \begin{tabular}{ |p{3cm}||p{3cm}|p{3cm}|p{3cm}|}
     \hline
     \multicolumn{4}{|c|}{CNN Feature Extraction Results} \\
     \hline
      & mAP & IoU & Inference time (ms) \\
     \hline
     SymboSLAM-Full &  43.6 &  0.6545 & 154 \\
     SymboSLAM-Tiny &  17.1 &  0.5877 & 842 \\
     \hline
    \end{tabular}
    
    }
\captionof{table}[CNN-Results]{Feature extraction results for Full size and tiny size of SymboSLAM CNN object detection modules. Results for full size obtained using NVIDIA GeForce GTX 1080 GPU with
Intel(R) Core(TM) i9-7900X CPU OS Linux. Results for tiny size obtained using CAT s64 Pro Android Smartphone Adreno 512 GPU Octa-core CPU OS Android 10}
\label{fig:CNN_Results}
\end{table}

\subsection{Symbolic Engine}\label{sec:Ontology_res}
The SymboSLAM ontology was run through the Ontology Pitfall Scanner,  OOPS!. Fourteen minor pitfalls are present within the ontology, but no critical or important pitfalls were detected. The SymboSLAM ontology was then used with the complete SymboSLAM architecture to conduct both simulated and real-world trials, as shown in~\ref{fig:gunghalin_comp} and~\ref{fig:civic_comp} (the remainder of simulated trials can be found in the appendix). The IoU and AP for the six simulated and two real-world trials are shown in Table~\ref{fig:SS_Results}. As the granularity of the 2D topographic environment maps generated by the SymboSLAM architecture could be segmented into a maximum of 24x24 partitions, the IoU and AP for the trials were calculated using all 24x24 segments for each generated environment map.

\begin{table}[t]
\normalsize
    \centering
    \resizebox{\textwidth/2}{!}{
   
    \begin{tabular}{ |p{4cm}||p{1.5cm}|p{1.5cm}|p{1.5cm}|p{1.5cm}|p{1.5cm}|}
     \hline
     \multicolumn{6}{|c|}{SymboSLAM Simulated and Measured Results} \\
     \hline
     \multirow{1}{*}{} &
     \multicolumn{2}{|c|}{Grid} &
      \multicolumn{2}{|c|}{Branch} &
      \multicolumn{1}{|c|}{} \\
     \hline
      & AP & IoU & AP & IoU & \# Features\\
     \hline
      
     Sim 01 - Gunghalin & 48.43 & 0.52 & 38.72 & 0.46 &  3694  \\
     Meas 01 - Gunghalin  & 8.51 &  0.27 & 1 & 0.01 & 2209 \\
     Sim 02 - Airport & 75.52 &  0.84  & 64.20 & 0.60 & 199\\
     Sim 03 - Fyshwick & 28.99 & 0.73 & 72.40 & 0.65 & 152  \\
     Sim 04 - Kingston & 33.85 & 0.90 & 41.20 & 0.75 & 212 \\
     Sim 05 - Train Depot & 8.16 & 0.90 & 31.99 & 0.31 & 167  \\
     Sim 06 - City  & 72.05 & 0.78 & 80.89 & 0.89 & 429 \\
     Meas 02 - City  & 19.27 &  0.48 & 1 & 0.01 & 833 \\
     \hline
      \multirow{1}{*}{Overall} &
        \multicolumn{2}{|c|}{AP} &
        \multicolumn{2}{|c|}{IoU} &
        \multicolumn{1}{|c|}{} \\
     \hline
     \multirow{1}{*}{Simulated} &
        \multicolumn{2}{|c|}{49.7} &
        \multicolumn{2}{|c|}{0.69} &
        \multicolumn{1}{|c|}{} \\
        
    \multirow{1}{*}{Measured} &
        \multicolumn{2}{|c|}{7.4} &
        \multicolumn{2}{|c|}{0.19} &
        \multicolumn{1}{|c|}{} \\
    \hline
    \end{tabular}
    
    }
\captionof{table}[CNN-Results]{IoU and AP for Full SymboSLAM architecture to generate 2D topo maps of environment types. Simulated results utilised an oracle system, and measured results utilised GPS.}
\label{fig:SS_Results}
\end{table}

\begin{figure*}[ht]
    \includegraphics[width=\textwidth]{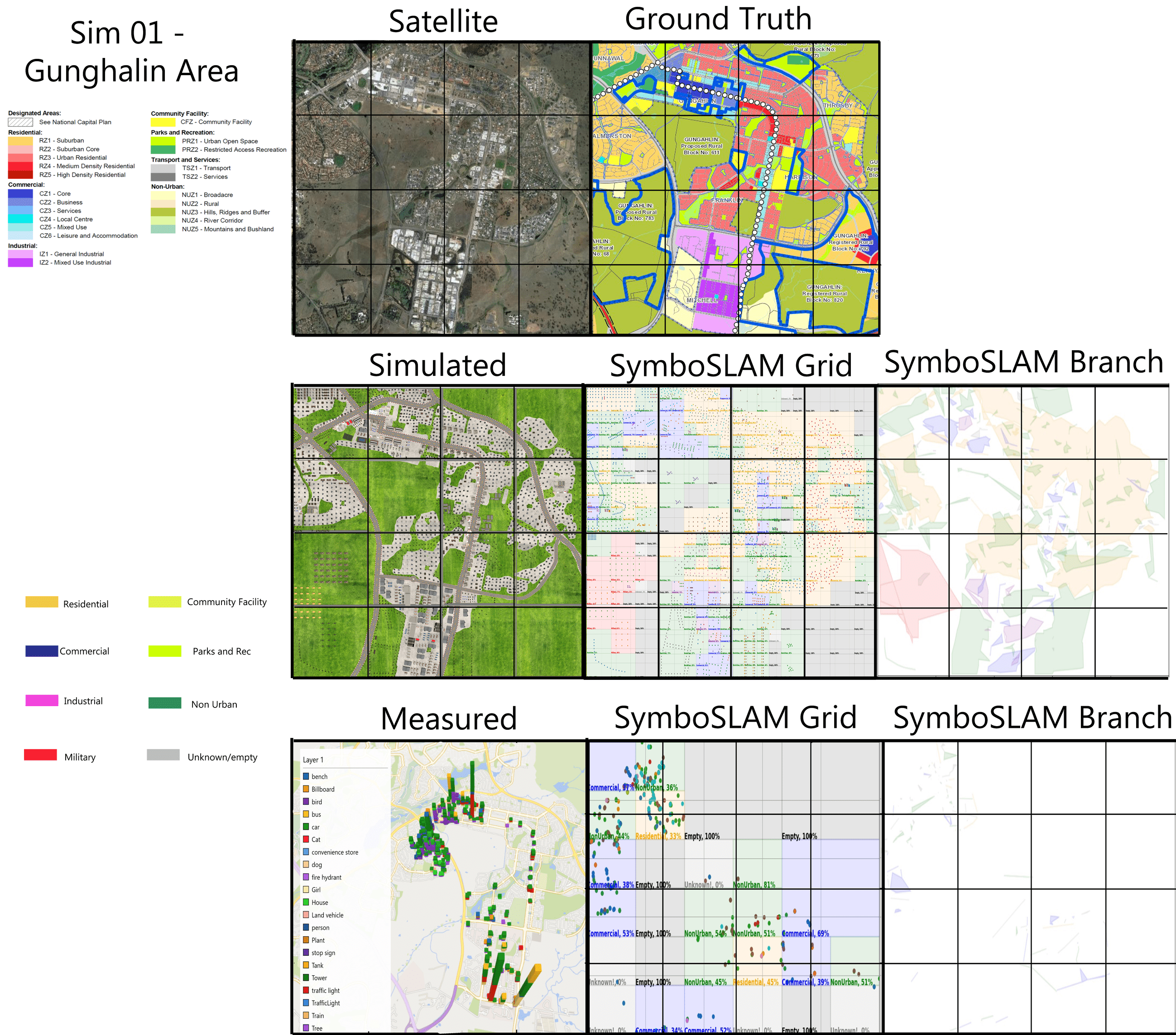}
    \caption{Comparison of ground truth and SymboSLAM results for ACT Gunghalin simulated and real-world area}
    \label{fig:gunghalin_comp}
\end{figure*}

\begin{figure*}[ht]
    \includegraphics[width=\textwidth]{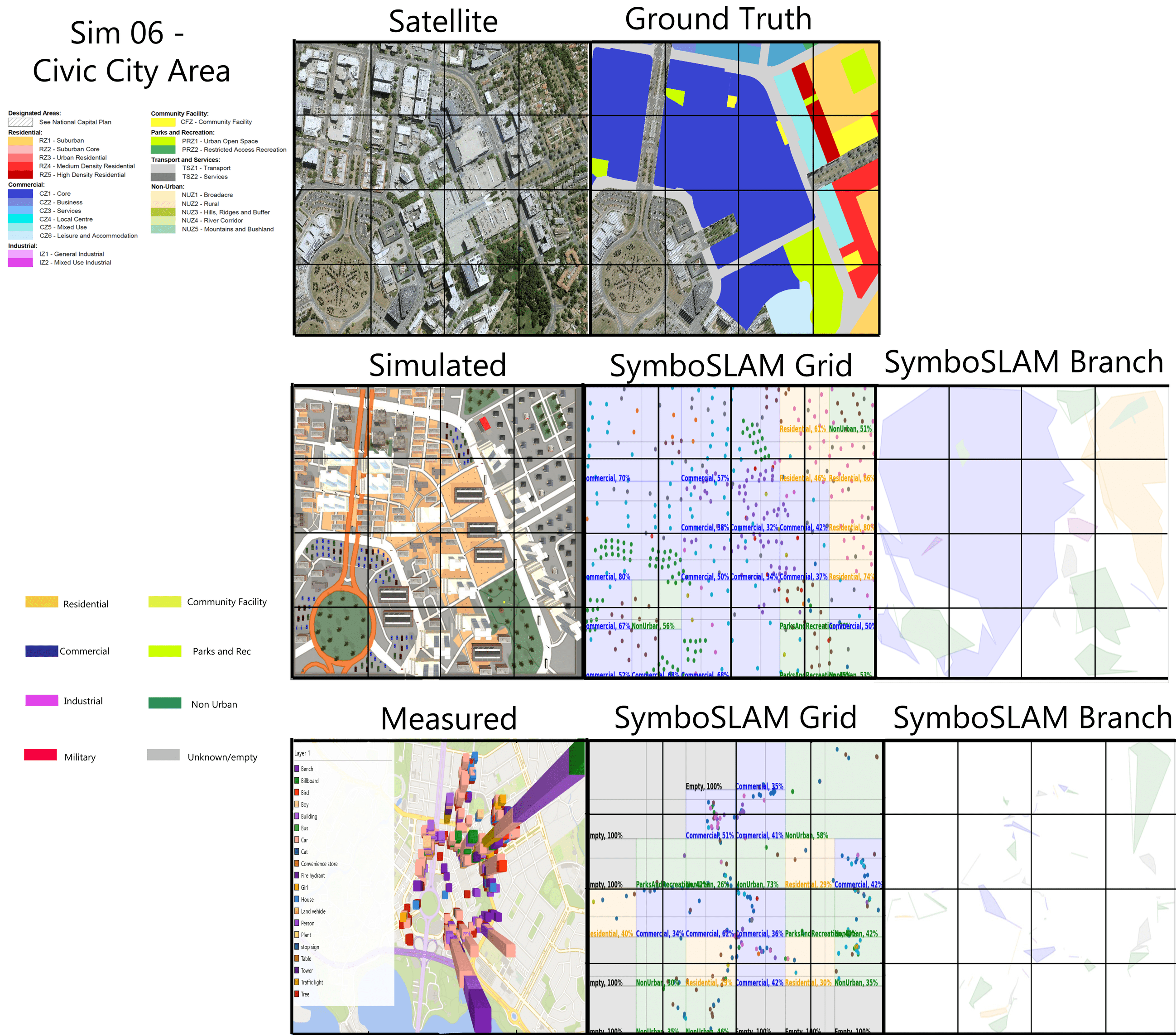}
    \caption{Comparison of ground truth and SymboSLAM results for ACT Civic simulated and real-world area}.
    \label{fig:civic_comp}
\end{figure*}

\section {Discussion}\label{sec:Discussion}

\subsection{Multi-Agent System} \label{sec:MAS_Disc}
Table~\ref{fig:MAS_Results} shows that simulations with a greater level of complexity and a higher number of features within the environment required a higher percentage of area coverage from the entire space available. This phenomenon is most noticeable in the area coverage difference between the Airport and City simulation, wherein an area of $\sim$ 74\% was required for seven items per square km for the Airport, and  $\sim$ 98\% was required for 238 items per square km for the city. Hence, the quasi-random search strategy is most effective for sparsely populated environments. Still, more complex environments would likely benefit from implementing more thorough search strategies such as a simple coordinated grid method. The time to finish, and normal dispersion, are a reflection of both simulation area size and complexity. The results from Table~\ref{fig:MAS_Results} describing this are hardly surprising (larger area leads to more time and greater dispersion), but this does demonstrate that the long-range coordinated targeting strategies implemented are functioning as intended.

\subsection{Simultaneous Localisation and Mapping} \label{sec:SLAM_Disc}
The trained YOLOV4 module used for feature extraction achieved a mAP of 14.3\% and 6.6\% lower than the scores achieved by the same network when trained on the MS COCO datasets~\cite{Bochkovskiy2020YOLOv4OS}~\footnote{https://pjreddie.com/darknet/yolo/}. The reduction in mAP and IoU is due to the reduction in images within the training dataset (328k in MS COCO to 200K in SymboSLAM). Other factors that may have affected the mAP and IoU could have included dissimilar feature types causing an underfit for low-level filters within the model, and the reduced number of feature classes reducing the complexity of the model. Updating the CNN feature extractor \footnote{YOLOv7 was released when this article was written - see https://paperswithcode.com/sota/real-time-object-detection-on-coco  for the latest models available} would likely improve the mAP and IoU of the module; however, the results achieved with the custom dataset were sufficient for the SymboSLAM architecture. Figure~\ref{fig:pred_output_sim} demonstrates the effectiveness of the trained feature extractor model in both simulation and the real-world environment.

The average error of centre offset for landmarks detected by edge agents, shown in Figure~\ref{fig:avg_error}, quite clearly indicate that the SLAM component of the SymboSLAM architecture does not yet function as intended. The recorded data shows a linear relationship between increased error and simulation time due to the compounding nature of the error in location measurement. The maximum recorded error in distance between landmark true and believed the location was almost 1.4 km, which is well out of tolerance for the SymboSLAM system requirements. The observed failure of the SLAM module is due to the error present in the Unity (and, by extension, EyeSim) sensors such as ImU and ToF.  Compounding location error is a problem-set present in the current literature for robotic SLAM systems~\cite{DBLP:journals/corr/abs-2106-10458}. At the outset of the SymboSLAM project, it was hypothesised that this problem would be negligible, as the granularity for feature extraction was much larger than most SLAM systems currently available, which was, however, not the case.

As such, to ensure the system's functionality and allow testing of other components, an oracle system was utilised to reveal landmarks on the map for which edge agents had discovered that were within a tolerable distance threshold to the true location of the landmark. The location of this landmark was then utilised by the oracle system to essentially 'reset' the average error of centre offset for that edge agent's map by adjusting the edge agent's landmark map to align better with this known location. Utilising a simulator / real sensors with a much lower measurement error will alleviate this issue for small-range areas (less than 1 square km) that are not complex (less than ten items per square km) but are not a viable long-term solution for the system. Many architectures offer loop closure as a solution for SLAM error issues~\cite{OrbSlam2, 3D_semantic_SLAM}. Notwithstanding, studies into this area have not yet generated a system capable of amending maps on a scale posed for the SymboSLAM problem. GPS offers a viable solution to this issue for both complex and large areas, as demonstrated when the system was deployed in the real environment in experiments 1-2. Relying on a sensor such as GPS for the SLAM module of the SymboSLAM system does, however, heavily limit the range of applications for which the system can be applied. Utilising the SymboSLAM architecture with GPS on a scale conducted in the eight trials will produce viable results. 

\subsection{Symbolic Engine} \label{sec:Symbolic_Eng_Disc}
An AP and IoU of 49.7 and 0.57 for the 12 simulated trials and 7.4 and 0.19 for the measured trials were achieved for the environment type classification utilising the SymboSLAM system. From Table~\ref{fig:SS_Results}, a clear relationship between the AP and the number of features is observed across the simulated results in that more features generate a higher AP of environment classification. An outlier to this relationship is the Airport which achieved the highest AP due to the ground truth of this area consisting of primarily non-urban areas, which is likely an oversight by the ACT government planning commission. By comparison, the sparsely populated area contained fewer community facility features than non-urban features, i.e. the sim had more trees than planes which highlight a shortcoming with the symbolic component of the SymboSLAM system - it does not take into account feature importance. When determining the classification of an area, some features are more important than others; for example, a skyscraper is indicative of only a city environment, whereas a tree could lead to many environments, as it is common in many areas. 

Table~\ref{fig:SS_Results} also shows that the branch segmentation method outperformed the grid segmentation method for all trials except for Gunghalin due to the increased white space (unknown area) introduced by the branch segmentation method. This white space is present in all trials but is most noticeable across the two measured trials in the Gunghalin simulated trial. Whilst this may be seen as a downfall of the system, it can also be observed as an advantage, as the branch segmentation method does not make assumptions across areas for which edge agents have not yet explored or for which there are no observable features. Theoretically, the IoU obtained from the branch segmentation method should be lower than the grid segmentation method, as the rigidity of the boundaries for this method is much more flexible. However, a higher IoU when using the branch segmentation method is not observed within the results of Table~\ref{fig:SS_Results} except for the city simulated trial. Feature abundance is again the reason for a decreased IoU within the branch segmentation method, as the boundaries between environment types are not clearly defined throughout the remainder of the simulated environments. The AP and IoU of all measured experiments were substantially lower than the simulated counterpart across both the grid and branch segmentation methods. 

Figures~\ref{fig:gunghalin_comp} and~\ref{fig:civic_comp} show a tendency for features to be clustered around streets and walkways due to the increased complexity involved with feature extraction from real-world environments. Space is present between all the features within the simulated environment as features extracted by the edge agents were placed on the control agents map at a level of homogeneity not achievable in the real world. Unknown environments are also much more prevalent within the measured results due to restricted and limited access areas hindering data collection and are most prevalent in the branch segmentation trials for the measured data collected. Hence,  the IoU and AP obtained will increase with increasing feature abundance and perform best with the grid segmentation method. This increased feature abundance can be easily achieved by altering the simulated models but would likely require another platform for data collection for measured results. Mounting the current platform to an aerial vehicle will likely alleviate this issue as it would obtain higher freedom of movement throughout the environment and, as such, would be able to collect data more homogeneously than the current method.

\section{Future Work} \label{sec:fut_work}

\subsection{Landmark Representation} \label{sec:landmark_rep}
A more accurate representation of landmarks should be incorporated into later iterations of the SymboSLAM architecture. Research looking to further this area may incorporate more traditional methods for SLAM to achieve this.

\subsection{Human Level Representation of Areas} \label{sec:abstract_map}
Incorporating the ability to understand human navigational aids such as signs~\cite{9091567} will increase the feature extraction ability of the SymboSLAM architecture and allow for a much richer depth of information to be accumulated about environmental features. For example, the SymboSLAM feature extraction modules can presently determine if an object is a building, but distinguishing between the types of buildings within a city is a task achievable through the understanding of signage.

\subsection{Simultaneous Localisation and Mapping} \label{sec:SLAM_fut_work}
The issues identified above within the SLAM modules of the SymboSLAM architecture must be addressed, and the solutions presently being investigated throughout the SLAM literature do not offer a viable solution for the scale of the problem posed in this paper. However, exploring the current oracle modules utilised for the SymboSLAM architecture may solve the SLAM problem. For example, a solution wherein landmarks are used in a resection formation to enable proper landmark belief generation, and subsequent confirmation through the map-matching process may be a workable solution.

\subsection{Hierarchical Chaining} \label{sec:Hierarchical_Chaining}
Building a belief generation of areas at a more granular level to be chained hierarchically may assist with improving the robustness and reliability. Increased explainability through an extended hierarchical chaining process may also enhance the trustworthiness of classified areas. 

\section{Conclusion}\label{sec:conclusion}
This paper proposes a novel approach to symbolic SLAM that uses symbolic reasoning through ontological design to create 2D environment-type maps through a multi-agent system with a hybrid orientation for contextual reasoning. The proposed system makes use of an intelligent edge agent architecture as well as a control agent architecture. Edge agents conduct local SLAM tasks within the environment through a random walk search strategy that extends the random-tree search method to introduce reactionary and long-range coordinated targeting. Intelligent edge agents semantically label extracted features and localise them spatially through place recognition, transforming the observed environment into a queryable set of state space representations. A control agent then receives many edge agents' maps to amend these maps (given the edge agent starting position) by conducting map matching techniques through a landmark map matching methodology to create a semantically labelled map of the environment. The semantics engine then takes in segmented partitions of these maps and utilises a purpose-built ontology to generate a probability distribution of likely environments. The SymboSLAM architecture was deployed in the Canberra region's simulation and the real world. It achieved an average precision and input over union of 49.7 and 0.57 for the 12 simulated trials and 7.4 and 0.19 for the measured trials, respectively, for the environment type classification.

\clearpage
\bibliographystyle{IEEEtran}
\bibliography{references}

\onecolumn
\clearpage

\begin{appendices}

\nomenclature{\(O\)}{Ontology}
\nomenclature{\(C\)}{Concepts}
\nomenclature{\(R\)}{Relationships}
\nomenclature{\(a\)}{Attributes}
\nomenclature{\(I\)}{Instances}
\nomenclature{\(A\)}{Axioms}
\nomenclature{\(SP\)}{Semantic Proximity}
\nomenclature{\(e_n\)}{Environment $n$}
\nomenclature{\(f_x, f_y\)}{Confidence of Feature $x$ and $y$}
\nomenclature{\(d\)}{Distance}
\nomenclature{\(C(e_n)\)}{Confidence of environment N}
\nomenclature{\(P(e_n)\)}{Probability of environment $n$}
\nomenclature{\(z\)}{Number of Features within a Segment}
\nomenclature{\(M\)}{Map}
\nomenclature{\(L\)}{Landmark}
\nomenclature{\(\mathbb{R}^D\)}{Vector space R with D dimensions}
\nomenclature{\(F\)}{Map segment}
\nomenclature{\( F_\text{env}\)}{Segment environment-type classification}
\nomenclature{\(L_\text{a1}\)}{Landmark 1 from map segment $a$}
\nomenclature{\(\mathbb{R}^M\)}{Vector space R with M dimensions}
\nomenclature{\(H\)}{Decision boundary hyper plane between two segments}
\nomenclature{\(AP\)}{Average Precision}
\nomenclature{\(mAP\)}{mean Average Precision}
\nomenclature{\(IoU\)}{Input over Union}
\nomenclature{\(\phi\)}{Mapping fucntion from input space to feature space}
\nomenclature{\(b\)}{Intercept and bias term of the SVM hyperplane equation in M dimensional space}
\nomenclature{\(w^T\)}{Gradient of decision boundary hyperplane}
\nomenclature{\(d_H(\phi(L_0))\)}{The distance between a SVM decision boundary hyperplane and a landmark point vector ($\phi(L_0)$)}
\nomenclature{\(||W||_2\)}{Euclidean norm for the length of w}
\nomenclature{\(W*\)}{Decision boundary updated in order to maximise distance between two landmark segment clusters}
\nomenclature{\(X\)}{Classification bounding box}
\nomenclature{\(Y\)}{Ground truth bounding box}
\nomenclature{\(TP\)}{True positive}
\nomenclature{\(FP\)}{False positive}
\nomenclature{\(FN\)}{False negative}
\nomenclature{\(pr\)}{Precision-Recall curve}
\nomenclature{\(Er\)}{Average error of centre offset}
\nomenclature{\(\)}{}


\printnomenclature

\end{appendices}

\end{document}